\documentclass[10pt]{article}
\newif\ifblindreview
\pdfoutput=1

\usepackage[letterpaper]{geometry}
\usepackage{amta2024}
\usepackage{times}
\usepackage{url}
\usepackage{latexsym}
\usepackage{natbib}
\usepackage{layout}
\usepackage{multicol}
\usepackage{booktabs,array}
\usepackage{float}
\usepackage[hidelinks]{hyperref}
\usepackage[all]{hypcap}
\usepackage{graphicx}
\usepackage{inconsolata}
\usepackage{xcolor}
\hypersetup{
    colorlinks,
    linkcolor={red!50!black},
    citecolor={blue!50!black},
    urlcolor={blue!80!black}
}

\usepackage[T1]{fontenc}
\usepackage[utf8]{inputenc}
\usepackage{anyfontsize}
\usepackage{amsmath}
\usepackage{amssymb}
\usepackage{amsfonts}
\usepackage{cgloss4e}
\usepackage{microtype}
\usepackage{newtxtext,newtxmath}

\usepackage{pgfplots}
\pgfplotsset{compat=1.18}
\usepackage{tikz}
\usetikzlibrary{positioning}

\newcommand{\at}{\scalebox{0.8}{@@}}
\newcommand{\atsp}{\scalebox{0.8}{@@}\hspace{0.33em}}

\usepackage{enumitem}
\newlist{compactitem}{itemize}{3}
\setlist[compactitem]{topsep=0pt,partopsep=0pt,itemsep=0pt,parsep=0pt}
\setlist[compactitem,1]{label=\textbullet}
\setlist[compactitem,2]{label=---}
\setlist[compactitem,3]{label=*}
\newlist{compactdesc}{description}{3}
\setlist[compactdesc]{topsep=0pt,partopsep=0pt,itemsep=0pt,parsep=0pt}
\newlist{compactenum}{enumerate}{3}
\setlist[compactenum]{topsep=0pt,partopsep=0pt,itemsep=0pt,parsep=0pt}
\setlist[compactenum,1]{label=\arabic*.}
\setlist[compactenum,2]{label=(\alph*)}
\setlist[compactenum,3]{label=\roman*.}

\ifblindreview
  \newcommand{\authorinfo}{\author{}}
\else
  \newcommand{\authorinfo}{
    \author{\name{\bf Kenneth J. Sible} \hfill  \addr{ksible@nd.edu}\\
            \name{\bf David Chiang} \hfill \addr{dchiang@nd.edu}\\
            \addr{\small Department of Computer Science and Engineering, University of Notre Dame, Notre Dame, 46556, United States}
    }
  }
\fi

\begin{document}

\twocolumn

\amtaHeader{x}{x}{xxx-xxx}{2015}{}{KJ Sible, D Chiang}
\title{\bf Improving Rare Word Translation \\ With Dictionaries and Attention Masking}
\authorinfo

\pagestyle{empty}

\twocolumn[
\vspace*{0.13in}
\maketitle
\begin{abstract}
\vspace*{10pt}
  In machine translation, rare words continue to be a problem for the dominant encoder-decoder architecture, especially in low-resource and out-of-domain translation settings. Human translators solve this problem with monolingual or bilingual dictionaries. In this paper, we propose appending definitions from a bilingual dictionary to source sentences and using \emph{attention masking} to link together rare words with their definitions. We find that including definitions for rare words improves performance by up to 1.0 BLEU and 1.6 MacroF1.
\end{abstract}
\vspace*{0.13in}
]

\section{Introduction}

The current state-of-the-art for machine translation (MT) is still the transformer encoder-decoder architecture \citep{kocmi_findings_2023}. While large language models such as LLaMA and GPT-4 have achieved great success on various NLP tasks, they still fall behind dedicated encoder-decoders for MT \citep{xuParadigmShiftMachine2023}. A major drawback of encoder-decoder models, however, is that they continue to struggle with rare word translation \citep{minh-congSimpleFastStrategy2022}.

Dictionaries, both monolingual and bilingual, are an indispensable resource for human translators, and in pre-neural statistical MT systems, it was common to use bilingual dictionaries to improve translation of rare words \citep{tan_passive_2015}. However, the use of dictionaries in neural MT is not straightforward, as there is a strong dependence on the surrounding context and word frequency in the training data \citep{wu2021taking}. In this paper, we explore a new approach for incorporating dictionaries into neural MT systems. We hypothesize that dictionaries could be useful both for low-resource translation, where the target language has limited training data, and out-of-domain translation, where the testing domain differs significantly from the training domain(s). In addition, dictionaries could facilitate continual learning by enabling zero-shot adaptation of MT systems.

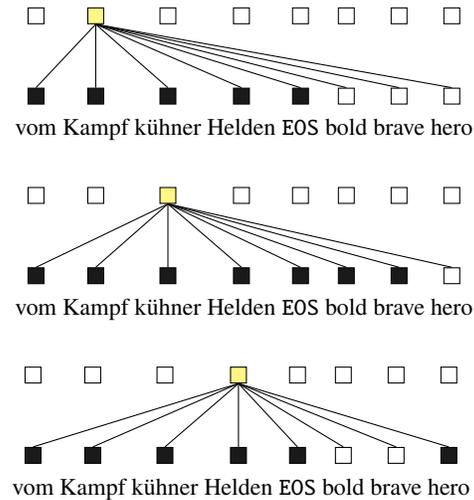
\begin{figure}[!ht]
    \centering\small

\bigskip

\begin{tikzpicture}
\begin{scope}[inner xsep=0pt,outer xsep=0pt,node distance=0.25em,font={\strut}]
\node (vom) {vom};
\node[right=of vom] (kampf) {Kampf};
\node[right=of kampf] (kuhner) {kühner};
\node[right=of kuhner] (helden) {Helden};
\node[right=of helden] (eos) {\texttt{EOS}};
\node[right=of eos] (bold) {bold};
\node[right=of bold] (brave) {brave};
\node[right=of brave] (hero) {hero};
\end{scope}
\begin{scope}[every node/.style={rectangle,draw,minimum size=6pt},node distance=0pt]
\begin{scope}[every node/.append style={fill=black!90}]
\node[above=of vom] (vom1) {};
\node[above=of kampf] (kampf1) {};
\node[above=of kuhner] (kuhner1) {};
\node[above=of helden] (helden1) {};
\node[above=of eos] (eos1) {};
\end{scope}
\begin{scope}[every node/.append style={fill=black!0}]
\node[above=of brave] (brave1) {};
\node[above=of bold] (bold1) {};
\node[above=of hero] (hero1) {};
\end{scope}
\end{scope}
\begin{scope}[every node/.style={rectangle,draw,minimum size=6pt},node distance=24pt]
\node[above=of vom1] (vom2) {};
\node[above=of kampf1,fill=yellow!60] (kampf2) {};
\node[above=of kuhner1] (kuhner2) {};
\node[above=of helden1] (helden2) {};
\node[above=of eos1] (eos2) {};
\node[above=of bold1] (bold2) {};
\node[above=of brave1] (brave2) {};
\node[above=of hero1] (hero2) {};
\end{scope}
\draw (kampf2.south) edge (vom1.north) edge (kampf1.north) edge (kuhner1.north) edge (helden1.north) edge (eos1.north) edge (bold1.north) edge (brave1.north) edge (hero1.north);
\end{tikzpicture}

\bigskip

\begin{tikzpicture}
\begin{scope}[inner xsep=0pt,outer xsep=0pt,node distance=0.25em,font={\strut}]
\node (vom) {vom};
\node[right=of vom] (kampf) {Kampf};
\node[right=of kampf] (kuhner) {kühner};
\node[right=of kuhner] (helden) {Helden};
\node[right=of helden] (eos) {\texttt{EOS}};
\node[right=of eos] (bold) {bold};
\node[right=of bold] (brave) {brave};
\node[right=of brave] (hero) {hero};
\end{scope}
\begin{scope}[every node/.style={rectangle,draw,minimum size=6pt},node distance=0pt]
\begin{scope}[every node/.append style={fill=black!90}]
\node[above=of vom] (vom1) {};
\node[above=of kampf] (kampf1) {};
\node[above=of kuhner] (kuhner1) {};
\node[above=of helden] (helden1) {};
\node[above=of eos] (eos1) {};
\node[above=of brave] (brave1) {};
\node[above=of bold] (bold1) {};
\end{scope}
\begin{scope}[every node/.append style={fill=black!0}]
\node[above=of hero] (hero1) {};
\end{scope}
\end{scope}
\begin{scope}[every node/.style={rectangle,draw,minimum size=6pt},node distance=24pt]
\node[above=of vom1] (vom2) {};
\node[above=of kampf1] (kampf2) {};
\node[above=of kuhner1,fill=yellow!60] (kuhner2) {};
\node[above=of helden1] (helden2) {};
\node[above=of eos1] (eos2) {};
\node[above=of bold1] (bold2) {};
\node[above=of brave1] (brave2) {};
\node[above=of hero1] (hero2) {};
\end{scope}
\draw (kuhner2.south) edge (vom1.north) edge (kampf1.north) edge (kuhner1.north) edge (helden1.north) edge (eos1.north) edge (bold1.north) edge (brave1.north) edge (hero1.north);
\end{tikzpicture}

\bigskip

\begin{tikzpicture}
\begin{scope}[inner xsep=0pt,outer xsep=0pt,node distance=0.25em,font={\strut}]
\node (vom) {vom};
\node[right=of vom] (kampf) {Kampf};
\node[right=of kampf] (kuhner) {kühner};
\node[right=of kuhner] (helden) {Helden};
\node[right=of helden] (eos) {\texttt{EOS}};
\node[right=of eos] (bold) {bold};
\node[right=of bold] (brave) {brave};
\node[right=of brave] (hero) {hero};
\end{scope}
\begin{scope}[every node/.style={rectangle,draw,minimum size=6pt},node distance=0pt]
\begin{scope}[every node/.append style={fill=black!90}]
\node[above=of vom] (vom1) {};
\node[above=of kampf] (kampf1) {};
\node[above=of kuhner] (kuhner1) {};
\node[above=of helden] (helden1) {};
\node[above=of eos] (eos1) {};
\node[above=of hero] (hero1) {};
\end{scope}
\begin{scope}[every node/.append style={fill=black!0}]
\node[above=of brave] (brave1) {};
\node[above=of bold] (bold1) {};
\end{scope}
\end{scope}
\begin{scope}[every node/.style={rectangle,draw,minimum size=6pt},node distance=24pt]
\node[above=of vom1] (vom2) {};
\node[above=of kampf1] (kampf2) {};
\node[above=of kuhner1] (kuhner2) {};
\node[above=of helden1,fill=yellow!60] (helden2) {};
\node[above=of eos1] (eos2) {};
\node[above=of bold1] (bold2) {};
\node[above=of brave1] (brave2) {};
\node[above=of hero1] (hero2) {};
\end{scope}
\draw (helden2.south) edge (vom1.north) edge (kampf1.north) edge (kuhner1.north) edge (helden1.north) edge (eos1.north) edge (bold1.north) edge (brave1.north) edge (hero1.north);
\end{tikzpicture}
    \caption{We append definitions of \emph{kühn} `bold, brave' and \emph{Held} `hero' to a sentence, and use an attention mask (with learnable strength) to inform the model which definitions correspond to which words. In each picture, the query vectors are above, with one query vector shaded yellow, and the key/value vectors are below, shaded to indicate the strength of the attention mask (black = not masked, white = masked).}
    \label{fig:attn-mask}
\end{figure}

However, the morphology of both the source and target language poses a major challenge for the use of dictionaries with MT compared to other NLP tasks that incorporate methods of retrieval-augmented generation \citep{niehuesContinuousLearningNeural2021}. MT systems that incorporate dictionaries must be capable of inflecting definitions for the target language and the context in which those definitions must appear, as dictionary entries and their definitions are often base forms. In Figure~\ref{fig:attn-mask}, we see the adjective \textit{kühner} is declined for the genitive case, but only the lemma \textit{kühn} would be in a German dictionary. Moreover, if the target language has adjective declension, then the MT system must also decline the dictionary form of the definition.

Our approach is to retrieve dictionary definitions for low-frequency words, append the definitions to source sentences containing rare words, and use attention masking to link together rare words with their definitions. We find that appending definitions for rare words improves MT performance by up to 1.0 BLEU and 1.6 MacroF1.

\section{Related Work}\label{related-work}

Previous work on dictionaries for neural MT can be divided into two broad categories, which we call \emph{dictionaries-as-translators} and \emph{dictionaries-as-text}. In the dictionaries-as-translators approach, the dictionary is assumed to contain high-quality translations of words, and the technical challenge is to get the MT system to use the dictionary's translations when appropriate. In the dictionaries-as-text approach, dictionary entries are added somehow to the source sentence, and it is up to the MT system to learn how to use them. In this approach, the dictionary can contain definitions that are not necessarily translations (e.g., one definition for German \emph{halt} is: ``Indicating that something is generally known, or cannot be changed, or the like; often untranslatable''). This approach could, in principle, use other resources like monolingual dictionaries, grammars, and so on.

\subsection{Dictionaries as translators}

In the dictionaries-as-translators category,
\citet{zhangPointDisambiguateCopy2021} propose a model with three steps: (1) identify source words that can be translated using a dictionary, (2) select one of several translation candidates (\textit{i.e.}, definitions), and (3) copy the selected translation into the output sequence. Similarly, other previous work in this category uses constrained decoding with a translation lexicon: \citet{zhang2016bridging}, \citet{arthur-etal-2016-incorporating}, \citet{fadaee-etal-2017-data}, \citet{chatterjee-etal-2017-guiding}, \citet{hasler-etal-2018-neural}, \citet{post-vilar-2018-fast}, \citet{thompson-etal-2019-hablex}, \citet{dinu-etal-2019-training}. 

A translation lexicon is a mapping of words from the source language to the target language, whereas a bilingual dictionary provides several possible translations for a given source word in addition to including definitions for untranslatable words such as particles. To incorporate a translation lexicon, we must constrain the output of the MT system, but that approach assumes the correct translation given the source context is contained within the lexicon. However, it quite often is the case that there are several valid translations with some being more appropriate than others for the given context.

\subsection{Dictionaries as text}

In the dictionaries-as-text category are approaches in which dictionary definitions are added to source sentences so that the model can learn how to use them. Two further questions arise: (1) How do we decide which definitions to include (especially in morphologically-rich languages, where a word in context does not in general match a dictionary headword)? (2) How do we represent the nonlinear structure of the input, which includes both a source sentence and associated definitions?

\citet{niehuesContinuousLearningNeural2021} lemmatizes each rare word and retrieves the matching bilingual definition, if any. The definition is inserted into the sentence immediately after the rare word, delimited by $\texttt{\#}$. He uses a combination of subword and character tokenization to improve handling of rare inflected forms.

\citet{zhongLookItBilingual2022} use a combination of Levenshtein distance and locality-sensitive hashing to find the closest dictionary headword for each, potentially inflected, rare word. They append the definitions to the end of the source sentence, and they inform the model about the structure of the input using position encodings (PEs). Each definition word's vector has contributions from both its own (sinusoidal) PE as well as the (learnable) PE of the defined word. They use BPE subword segmentation for all words; instead of the PE of the defined word, they choose the PE of its first subword. In contrast to \citet{niehuesContinuousLearningNeural2021}, \citet{zhongLookItBilingual2022} find that the model with BPE can inflect dictionary definitions without switching to character-level tokenization. 

\section{Methodology}\label{methodology}

Our approach falls squarely into the dictionaries-as-text category: given a source sentence, we retrieve relevant entries from a bilingual dictionary and include them in the source sentence. To decide which entries to include, we use a source-language lemmatizer, which should be more reliable and faster than fuzzy matching. To represent the input, we use \emph{attention masking} instead of positional encodings since we suspect that attention is a more natural mechanism by which an encoder-decoder model can associate definitions (keys) with rare words (queries).

In this section, we break down our approach for using a bilingual dictionary for machine translation with a transformer-based, encoder-decoder model into the following steps: (1) headword selection, (2) definition retrieval, and (3) attention masking.

\subsection{Headword Selection}
In order to classify a source word as \textit{rare}, we compare the number of occurrences in the training data against a frequency threshold that we choose from a hyperparameter search. For a given source word, we say that a word $w$ is \textit{rare} if (a) it has both a frequency below the threshold and an entry in the dictionary, or (b) if $w$ does not meet either of the above criteria, but its lemmatized form meets both.

\subsection{Definition Retrieval}
If a rare/unknown word is present in the dictionary, we retrieve its definition(s). Otherwise, we first use a lemmatizer and check if the dictionary contains the lemma for the rare/unknown word. Then, we append the definition(s) to the source sentence following the end-of-sentence token \verb!<EOS>!.

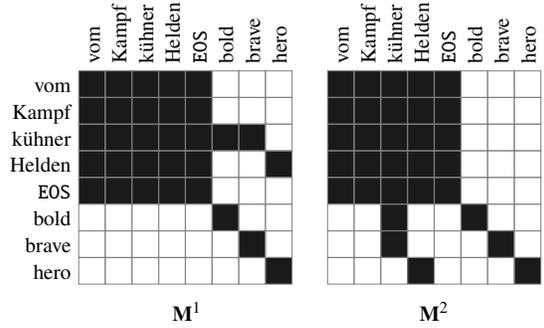
\begin{figure}[!ht]
    \newcommand{\bn}{\node[fill=black!90,draw=black!50,line width=0.1pt,minimum size=10pt]{};}
\newcommand{\wn}{\node[fill=white,draw=black!50,line width=0.1pt,minimum size=10pt]{};}
\newcommand{\hw}[1]{\node[anchor=east]{\smash{\raisebox{-3pt}{#1}}};}
\newcommand{\vw}[1]{\node[anchor=west,rotate=90]{\smash{\raisebox{-3pt}{#1}}};}

\centering\footnotesize

\begin{tabular}{@{}c@{\quad}c@{}}
\begin{tikzpicture}
\matrix {
& \vw{vom} & \vw{Kampf} & \vw{kühner} & \vw{Helden} & \vw{\texttt{EOS}} & \vw{bold} & \vw{brave} & \vw{hero} \\
\hw{vom} & \bn & \bn & \bn & \bn & \bn & \wn & \wn & \wn \\
\hw{Kampf} & \bn & \bn & \bn & \bn & \bn & \wn & \wn & \wn \\
\hw{kühner} & \bn & \bn & \bn & \bn & \bn & \bn & \bn & \wn \\
\hw{Helden} & \bn & \bn & \bn & \bn & \bn & \wn & \wn & \bn \\
\hw{\texttt{EOS}} & \bn & \bn & \bn & \bn & \bn & \wn & \wn & \wn \\
\hw{bold} & \wn & \wn & \wn & \wn & \wn & \bn & \wn & \wn \\
\hw{brave} & \wn & \wn & \wn & \wn & \wn & \wn & \bn & \wn \\
\hw{hero} & \wn & \wn & \wn & \wn & \wn & \wn & \wn & \bn \\
};
\end{tikzpicture}
&
\begin{tikzpicture}
\matrix {
\vw{vom} & \vw{Kampf} & \vw{kühner} & \vw{Helden} & \vw{\texttt{EOS}} & \vw{bold} & \vw{brave} & \vw{hero} \\
\bn & \bn & \bn & \bn & \bn & \wn & \wn & \wn \\
\bn & \bn & \bn & \bn & \bn & \wn & \wn & \wn \\
\bn & \bn & \bn & \bn & \bn & \wn & \wn & \wn \\
\bn & \bn & \bn & \bn & \bn & \wn & \wn & \wn \\
\bn & \bn & \bn & \bn & \bn & \wn & \wn & \wn \\
\wn & \wn & \bn & \wn & \wn & \bn & \wn & \wn \\
\wn & \wn & \bn & \wn & \wn & \wn & \bn & \wn \\
\wn & \wn & \wn & \bn & \wn & \wn & \wn & \bn \\
};
\end{tikzpicture}
\\
\hspace{3.5em} $\mathbf{M}^1$ & $\mathbf{M}^2$
\end{tabular}
    \caption{Our system uses two attention masks with learnable strengths. Rows are queries; columns are keys/values. Black = not masked; white = masked. Mask $\mathbf{M}^1$ allows each source word to attend to its definitions (if any). Mask $\mathbf{M}^2$ allows each definition word to attend to the word it defines.}
    \label{fig:mask}
\end{figure}

\subsection{Attention Masking}

The input now contains a source sentence augmented with dictionary definitions, both segmented into subwords using BPE. To inform the model about the structure of the input, we use attention masking \citep{tao_shen_disan_2018}. 

Let $n$ be the input length (source subwords plus definition subwords), and let $d$ be the dimensionality of the model's hidden vectors. In standard attention, we compute, for each head $h$, a matrix of attention weights $\alpha_h \in \mathbb{R}^{n \times n}$:
\begin{equation*}
    \alpha_h = \mathrm{softmax}\left(\frac{\mathbf{Q}_h\mathbf{K}^T_h}{\sqrt{d}}\right)
\end{equation*}
where $\mathbf{Q}_h, \mathbf{K}_h \in \mathbb{R}^{n \times d}$ are the query and key matrices, respectively, for head $h$.

We construct two masks (see Figure~\ref{fig:mask}). Both masks allow all source subwords to attend to all source subwords, and all definition subwords to attend to all subwords in the same definition. Note that \emph{kühner} has two definitions, which cannot attend to each other. 
Mask $\mathbf{M}^1$ allows each source subword to attend to its definitions (if any). Mask $\mathbf{M}^2$ allows each definition subword to attend to the word it defines. Mathematically, we represent each mask as a matrix $\mathbf{M}^k \in \{0,1\}^{n\times n}$, where $\mathbf{M}^k_{ij} = 1$ means that subword $i$ cannot attend to subword $j$.

The attention masks are applied softly, with learnable weights. We combine the masks as follows:
\begin{equation*}
    \alpha_h = \mathrm{softmax}\left(\frac{\mathbf{Q}_h\mathbf{K}_h^T}{\sqrt{d}}-\sum_{k=1}^{m}\exp\left(s_{k,h}\right)\mathbf{M}^k \right)
\end{equation*}
where $m=2$ is the number of masks and $s_{k,h} \in \mathbb{R}$ is the learnable strength for mask $k$ and head $h$. We apply the exponential function component-wise to each $s_{k,h}$ to ensure that every element of the summation is positive. The aggregate attention mask is then subtracted from the standard dot-product attention. In this way, the model can decide if and/or when the dictionary definitions are useful and adjust the strengths of the attention masks accordingly \citep{mcdonaldSyntaxBasedAttentionMasking2021}.

\section{Experiments}\label{experiments}

\begin{table*}[t]
\centering
\begin{tabular}{ll|cc|cc}
\toprule
&&\multicolumn{2}{c|}{\textbf{News}} & \multicolumn{2}{c}{\textbf{Biomedical}} \\
\textbf{Training Corpus} & \textbf{Model} & \textbf{BLEU} & \textbf{MacroF1} & \textbf{BLEU} & \textbf{MacroF1} \\
\midrule
Europarl (Limited) & Baseline & $22.1$ & $18.1$ & $18.2$ & $18.5$ \\
& Parallel & $22.3$ & $18.2$ & $18.2$ & $18.4$ \\
& DPE & $22.4$ & $18.4$ & $18.3$ & $18.6$ \\
& Masking & $\mathbf{23.4}$ & $\mathbf{20.0}$ & $\mathbf{19.1}$ & $\mathbf{19.9}$ \\
\midrule
Europarl (Full) & Baseline & $30.4$ & $25.4$ & $23.8$ & $25.8$ \\
& Parallel & $30.5$ & $25.4$ & $24.0$ & $25.9$ \\
& DPE & $31.1$ & $26.3$ & $24.3$ & $26.5$ \\
& Masking & $\mathbf{31.2}$ & $\mathbf{26.8}$ & $\mathbf{24.4}$ & $\mathbf{26.9}$ \\
\bottomrule
\end{tabular}
\quad
\begin{tabular}{cc}
\toprule

\midrule
\bottomrule
\end{tabular}
\caption{\label{tab:metrics}
Baseline refers to the translation model without any dictionaries, Parallel includes a bilingual dictionary as parallel text, DPE appends dictionary definitions and uses positional encodings \citep{zhongLookItBilingual2022}, and Masking (ours) appends dictionary definitions and uses attention masking. To construct Europarl (Limited), we only use the first 250,000 sentences (<10\%) of the 1.8 million in Europarl (Full).
}
\end{table*}

In this section, we describe our translation model, the source-side lemmatizer, and the bilingual dictionary used hereafter throughout the paper.

\subsection{Translation Model}
To experiment with the internal architecture, we implement an encoder-decoder model from scratch using \verb!PyTorch! \citep{paszke_pytorch_2019}.\footnote{\url{https://github.com/kennethsible/dictionary-attention}} For the encoder/decoder, we use a transformer model \citep{vaswani_attention_2017}. Hidden vectors have $d = d_{\text{model}} = 512$ dimensions, and feed-forward networks have $d_{\text{FFN}} = 2048$ dimensions. The encoder and decoder each have 6 layers, each with 8 attention heads. We apply dropout to all embedding, feed-forward, and attention layers with a probability of $0.1$. Instead of layer normalization, we use \verb!FixNorm! and \verb!ScaleNorm!, which have been shown to improve translations in the low-resource setting \citep{nguyenTransformersTearsImproving2019}.

All models are trained on NVIDIA A10 GPUs. We use negative log-likelihood for training with a batch size of 4096, a label smoothing value of 0.1, and an initial learning rate of $3 \cdot 10^{-4}$, which we decay by a factor of $0.8$ with a patience of $3$ and a minimum learning rate of $5 \cdot 10^{-5}$. In addition, we do early stopping if our model trains for 20 epochs without improvement or exceeds a maximum of 250 epochs. Finally, we filter through the training data by removing empty translations, duplicate sentence pairs, sentences longer than a maximum length of 256, and sentence pairs with a source:target length ratio greater than 1.3. We also normalize the punctuation in both the source and target languages.

\subsection{Training/Evaluation Data}

For German to English translation, we use data from the WMT22 shared task: Europarl v10 for training \citep{koehnEuroparlParallelCorpus2005}, newstest2019 for validation, and newstest2022 for testing. For tokenization, we use \verb!sacremoses!,\footnote{\url{https://github.com/hplt-project/sacremoses}} an implementation of Moses \citep{koehn_moses_2007}, at the word-level and \verb!subword-nmt!,\footnote{\url{https://github.com/rsennrich/subword-nmt}} an implementation of BPE, at the subword-level. For evaluation, we use a fork of \verb!sacrebleu! \citep{post_call_2018} for BLEU \citep{papineni_bleu_2002} and MacroF1 \citep{gowda_macro-average_2021}.\footnote{\url{https://github.com/isi-nlp/sacrebleu}}

The Europarl corpus for German to English has 1,778,520 sentences, with 1,379,973 remaining after cleaning. We apply BPE with 32,000 merge operations and a dropout probability of 0.1 to obtain a shared vocabulary size of 32,469. The newstest2019 validation set and newstest2022 test set contain 2,000 and 1,984 sentences, respectively. To measure translation performance in a low-resource setting, we limit the Europarl corpus to the first 250,000 sentences. The smaller training set has 190,686 sentences remaining after cleaning. We apply BPE with 8,000 merge operations and a dropout probability of 0.1 to obtain a shared vocabulary size of 8,348.

Regarding the difficulty of finding and/or curating extensive dictionaries for low-resource languages, the available Uyghur-English data for the DARPA LORELEI Year 1 evaluation \citep{hermjakobIncidentDrivenMachineTranslation2018}, for example, consisted of 99k sentences of parallel text and 240k dictionary entries, so there are cases where the amount of dictionary data available is extensive compared to the amount of parallel text available. Given the lack of available training data for low-resource languages, we would argue that hiring linguists to construct bilingual dictionaries offers a greater overall benefit to the community of native speakers and those wishing to document/preserve/revitalize the language than simply hiring translators to expand the available corpora, as the usefulness of dictionaries extends beyond NLP applications \citep{garrette-baldridge-2013-learning}.

To evaluate the performance of our model on out-of-domain translation, we combine the Medline test sets from the WMT20 \citep{bawden-etal-2020-findings}, WMT21 \citep{yeganova-etal-2021-findings}, and WMT22 \citep{neves-EtAl:2022:WMT} biomedical tasks, removing any duplicate sentence pairs. However, the parallel text is misaligned, so we use the provided alignment files to construct the test set, filtering out all sentence pairs not labeled as \verb!OK!. The final test set has 1,073 sentences.
Table~\ref{tab:biomedical} shows an example sentence from the biomedical test set along with the reference translation.

\subsection{Lemmatizer and Dictionary}
For the German lemmatizer, we used the spaCy model \verb!de_core_news_sm!\footnote{\url{https://spacy.io/models/de}} with only the tok2vec, tagger, and lemmatizer enabled in the NLP pipeline. For the bilingual dictionary, we used the most recent development version of the German to English bilingual dictionary provided by TU Chemnitz.\footnote{\url{https://ftp.tu-chemnitz.de/pub/Local/urz/ding/de-en-devel/}} To prepare the data for our model, we filtered out:
\begin{compactitem}
    \item All dictionary headwords labeled non-alphabetic in Python, excluding hyphenated compound (\textit{e.g.}, \texttt{im eigenen Tempo}).
    \item All dictionary metadata contained in grouping symbols, such as part-of-speech and gender (\textit{e.g.}, masculine noun \texttt{\{m\}}, transitive verb \texttt{\{vt\}}, biological term \texttt{[biol.]}, Austrian dialect \texttt{[Ös.]}).
    \item All dictionary abbreviations used for nominative, accusative, dative, and genitive objects (\textit{e.g.}, \texttt{jdm.}, \texttt{jdn.}, \texttt{jds.}, and \texttt{etw.}).
    \item All German prepositional phrases of the form: preposition + abbreviation (\textit{e.g.}, \texttt{bei jdm./etw.}).
    \item The German reflexive pronoun \textit{sich} whenever preceding a headword (\textit{e.g.}, \texttt{sich anschließen}).
\end{compactitem}

The German to English dictionary, after cleaning and applying the filters, has 302,061 entries.

\subsection{Experimental Setup}
In addition to our model (Masking), we trained three baseline models: a translation model without any dictionaries (Baseline), a model that includes a bilingual dictionary as parallel text (Parallel), and a model that uses dictionary positional encodings (DPE) \citep{zhongLookItBilingual2022}. For DPE and Masking, we append dictionary definitions to source sentences containing rare words. All models were trained on two datasets: Europarl (Limited) and Europarl (Full).

\subsection{Hyperparameter Search}
By appending dictionary definitions, we introduce two hyperparameters in the model: the frequency threshold for rare words and the number of definitions (or word senses) appended for each rare word. In our experiments, we used frequency thresholds of 5, 10, 15, 25, and 50, and restricted the number of definitions appended to 1, 5, 10, and unbounded.

\section{Results}\label{results}

{\newcommand{\gloss}[1]{\raisebox{12pt}{\parbox[t]{13cm}{\glll #1}}}
\renewcommand{\eachwordtwo}{\tt\scriptsize}
\renewcommand{\eachwordthree}{\scriptsize}

\newcommand{\child}{\rotatebox[origin=c]{180}{$\Lsh$}}
\renewcommand{\arraystretch}{1.1}

\begin{table*}
\centering \small \begin{tabular}{@{}l@{\hspace{0.5em}}p{13.7cm}@{}}
\toprule
Source & \gloss{Mit seiner \textbf{Tarnkappe} \textbf{entkam} \textbf{Siegfried} dem \textbf{kampferprobten} \textbf{Ritter}, einem \textbf{Todfeind}, und \textbf{schlich} sich aus der Burg. \\ Mit seiner {T\atsp ar\atsp n\atsp k\atsp app\atsp e} {ent\atsp kam} {Sie\atsp g\atsp fri\atsp ed} dem {kampf\atsp er\atsp prob\atsp ten} {R\atsp it\atsp ter ,} einem {To\atsp d\atsp fein\atsp d ,} und {sch\atsp lich} sich aus der {Burg .} \\ with his {invisibility cloak} evaded Sigurd the battle-hardened knight a {deadly enemy} and crept himself {out of} the castle \\} \\&
\textbf{Tarnkappe}: \{invisibility cloak\};
\textbf{entkam}: \{evaded, escaped, got away\};
\textbf{Siegfried}: \{Sigurd\};\\&
\textbf{kampferprobt}: \{battle-seasoned, battle-hardened, battle-tested, combat proven\};\\&
\textbf{Ritter}: \{knight, knights, companion of the order of knighthood, chevalier\};\\&
\textbf{Todfeind}: \{deadly enemy, mortal enemy\};
\textbf{schlich}: \{crept, slunk, tiptoed\}
\\
Reference & With his invisibility cloak, Siegfried evaded the battle-hardened knight,
a deadly foe, and crept out of the castle. \\
Baseline & With his cap, Siegfried escaped the tried and tested ritter,
a death-enemy, and smashed from the castle. \\
Parallel & With his cap, Siegfried escaped the tried and tested Ritter,
a death enemy, and came out of the castle shamefully. \\
DPE & With his glasscloak, Sigurd escaped the fighter's knight, a deadly enemy, and crept out of the castle. \\
Masking & With his invisibility cloak, Sigurd escaped the battle-tested knight,
a deadly enemy, and crept out of the castle. \\
Apple & With his camouflage cap, Siegfried escaped the battle-tested knight,
a mortal enemy, and crept out of the castle. \\
\bottomrule
\end{tabular}

\caption{
On a German sentence (Source), our system's output (Masking) is closer to the Reference than the Baseline system's, even when the dictionary is included in the baseline system's training data (Parallel) or dictionary positional encodings \citep{zhongLookItBilingual2022} are used instead of attention masking (DPE). Even Apple's Translate app translates \emph{Tarnkappe} over-literally as \emph{camouflage cap}.
Rare words are written in boldface.
The Reference sentence was written by the first author to demonstrate multiple rare words with a variety of parts of speech and inflections, and a native German speaker translated it into the Source sentence.}
\label{tab:nibelung}
\end{table*}

\begin{table*}
\centering \small
\begin{tabular}{@{}l@{\hspace{0.8em}}p{13cm}@{}}
\toprule
Source & \gloss{Typisch für ein \textbf{konjunktivales} \textbf{Lymphom} ist eine \textbf{lachsfarbene} \textbf{Schwellung}. \\ {Typ\atsp isch} für ein {kon\atsp jun\atsp ktiv\atsp ales} {L\atsp ymp\atsp ho\atsp m} ist eine {la\atsp chs\atsp far\atsp bene} {Schwell\atsp ung .} \\ typical for a conjunctival lymphoma is a salmon-colored swelling \\} \\&
\textbf{Lymphom}: \{lymphoma\};
\textbf{lachsfarben}: \{salmon, salmon-coloured, salmon-colored\};\\&
\textbf{Schwellung}: \{swelling-up, swelling, puffiness, tumescence, intumescence, intumescentia, tumentia, tumefaction, tumidity, turgescence, turgidity, engorgement\}
\\
Reference & A salmon-colored swelling is typical for conjunctival lymphoma.  \\
Baseline & A lax threshold is typical of a lax lymphom in economic terms.  \\
Parallel & A low level threshold is typical of a cyclical lymphom. \\
DPE & A cyclical lymphom is typically characterised by a lame threshold. \\
Masking & A salmon-coloured lymphoma is typical of a cyclical lymphoma.  \\
\child{} Restricted & A salmon-coloured swelling is typical of a current lymphoma. \\
\child{} Updated & A salmon-coloured swelling is typical of a conjunctival lymphoma. \\
\bottomrule
\end{tabular}
\caption{
On a German sentence (Source) from the biomedical dataset, our system's output (Masking) is closer to the Reference than the Baseline system's, even when the dictionary is included in the baseline system's training data (Parallel) or dictionary positional encodings \citep{zhongLookItBilingual2022} are used instead of attention masking (DPE).
Rare words are written in boldface.
We also edited the input manually for demonstration purposes: For Restricted, the number of definitions appended has been restricted to 3 since \emph{Schwellung} has 12, which causes the model to struggle. For Updated, we restricted the number of definitions appended to 3 and added a definition for \emph{konjunktival} `conjunctival' to the dictionary (not previously present).}
\label{tab:biomedical}
\end{table*}}

\begin{figure*}
    \input{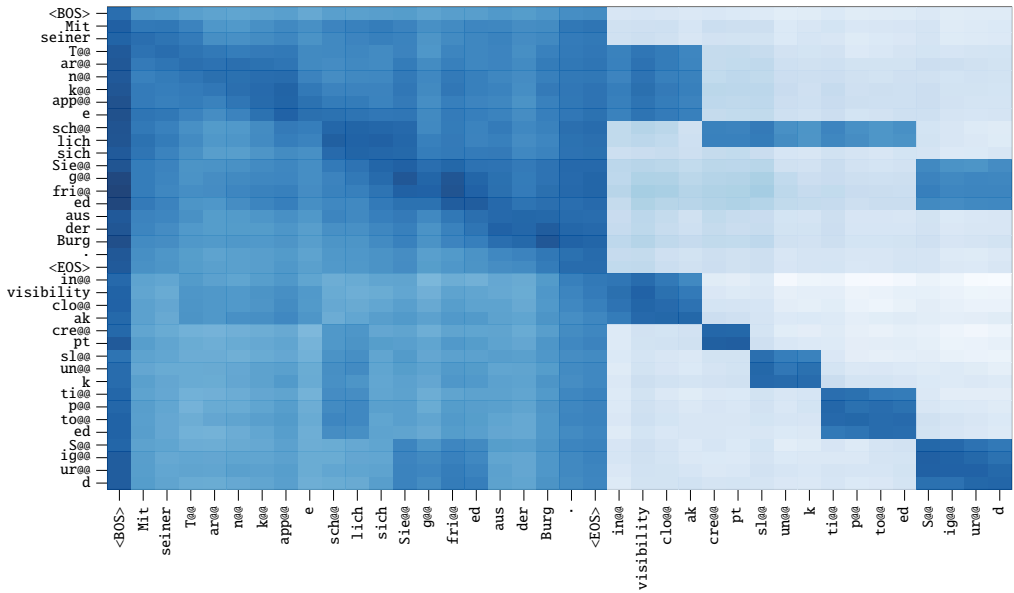}
    \caption{The attention scores of the Masking model for the German sentence: ``Mit seiner Tarnkappe schlich sich Siegfried aus der Burg'' with the definition string ``invisibility cloak crept slunk tiptoed Sigurd.'' The attention scores are summed for all encoder layers and attention heads. We observe both attention masks being utilized by the model.}
    \label{fig:attn-heat}
\end{figure*}

In this section, we report and analyze the results of our experiments described in the previous section. We found that appending definitions for rare words and using attention masking (Masking) improved translation performance over the baseline models: Baseline, Parallel, and DPE.

Furthermore, we observed that using a lower frequency threshold during training and increasing that threshold during inference resulted in the largest performance improvement. We speculate that this behavior is a result of larger thresholds incorrectly classifying unknown compound words (\textit{i.e.}, those not occurring in the training data) as rare when they are already translatable by the baseline models as a result of subword tokenization. If we are attempting to teach the model to use definitions, including them when they are not necessary may actively work against our training objective.

During a hyperparameter search, we found that using 10 for the frequency threshold and 10 for the maximum number of definitions appended yields the largest improvement in translation performance. Table~\ref{tab:metrics} compares our model against the three baseline models for both the general (news) and out-of-domain (biomedical) datasets. All metrics reported in the table were averaged over 5 random restarts and statistical significance was verified with paired t-tests. Masking (our model) outperforms the three baseline models on both metrics and all improvements are statistically significant (p-value < 0.05), except for the BLEU improvement over DPE for Europarl (Full).

In the low-resource setting, DPE struggles to improve over the baseline models, while Masking has the largest BLEU improvement, demonstrating a boost in low-resource translation performance. In the high-resource setting, although DPE and Masking are not significantly different in BLEU, they are significantly different in MacroF1. Since rare words are less frequent in the high-resource setting, the BLEU improvement of DPE and Masking over Baseline and Parallel is not as large. However, Masking has the largest MacroF1 improvement, demonstrating a boost in rare word translation performance.

In Table~\ref{tab:nibelung}, we compare candidate translations of a German sentence containing rare words against an English reference. The sentence was written in English, and translated by a native German speaker, to demonstrate the capability and robustness of our model in using the dictionary. The German sentence contains seven rare words of varying part-of-speech, including adjective declension and verb conjugation. In the Source row, the English glosses are shown beneath each German word to match the Reference translation along with the corresponding subword tokenization. To reduce the sentence length, the definitions are listed separately instead of appended to the German sentence.

Baseline and Parallel contain several incorrect translations of rare words. In particular, we observe that \textit{Tarnkappe} and \textit{Todfeind} were translated over-literally, with the first noun in the compound being dropped all together. Even the Apple Translate app translated \textit{Tarnkappe} over-literally as \textit{camouflage cap}. The DPE model, instead of dropping the first noun like Baseline/Parallel, used a seemingly random noun and translated the second over-literally. Only the Masking model correctly translated \textit{Tarnkappe} as \textit{invisibility cloak}. In fact, Masking used at least one definition for every rare word, getting the closest to the Reference.

In Table~\ref{tab:biomedical}, we compare candidate translations of a German sentence taken directly from the Medline test set. The sentence contains four rare words, but our dictionary has no definition for \textit{konjunktival}. \textit{Lymphom}, despite having a definition, is copied to English sentence by Baseline, Parallel, and DPE. Masking correctly translates \textit{Lymphom} and \textit{lachsfarben}, but all models mistranslate \textit{Schwellung}. We found that Masking often ignores definitions if there are too many appended for a given rare word. To demonstrate, we restricted the number of definitions for \textit{Schwellung} to 3 and see that the model correctly translates the word. We also succeeded in translating \textit{konjunktival} correctly by adding the English definition to the dictionary, demonstrating that the dictionary coverage is a limiting factor.

In Figure~\ref{fig:attn-heat}, we use an attention heat map to visualize the attention scores for a German sentence. The sentence shown is a trimmed version of the example in Table~\ref{tab:nibelung}. To build the heat map, we summed the attention scores for every encoder layer and every attention head. We see that the attention masks shown in Figure~\ref{fig:mask} are clearly visible in the heat map. However, the model decided to put more emphasis on the first mask than the second, which is done by adjusting the mask strengths.

\section{Discussion}

\paragraph{Rare Word Classification}
As mentioned previously, compound words that do not occur in the training data may still be accurately translated as a result of subword segmentation, suggesting that frequency is not an ideal or reliable metric for classifying rare words. In the future, frequency could be replaced with a source-side estimation of model confidence in the translation of rare words. 

\paragraph{Incorrect Lemmatization}
We could not find an acceptable lemmatizer for the German language  since even spaCy would occasionally misidentify the lemma for, \textit{e.g.}, a declined adjective or a past participle. Furthermore, no lemmatizer that we found could correctly identify the infinitive form for separable verbs or \textit{trennbare Verben}, a common class of verbs in the German language. In the future, we could explore more robust lemmatization techniques or the inclusion of inflected forms in the dictionary.

\paragraph{Lemmatization Ambiguities}
We have identified several cases where lemmatization causes the model to use a definition that is not grammatically correct in the context of the source sentence. For example, if the past tense form of a verb is not present in the dictionary and the definition for the infinitive form is used, the model often avoids inflecting the infinitive form to the correct tense unless the sentence contains, \textit{e.g.}, an auxiliary verb. Similarly, nouns ending in --er in German have no plural ending, which creates an ambiguity as to whether the English definition should be plural. A dictionary that directly contains inflected forms may resolve such ambiguities.

\paragraph{Definition/Word Sense Pruning}
We appended definitions for each word sense and part-of-speech with the assumption that the model could learn to leverage syntactic or semantic knowledge of the source sentence to select an appropriate translation for the rare words from among those definitions appended. However, we find that the model is often spoiled for choice, in that the model may use an inappropriate definition, or none at all, if there are too many without a clear way to disambiguate. In the future, we could implement a strategy to select the most relevant definitions or limit the number of appended definitions per rare word, such as pruning based on document-level context or prior domain knowledge.

\paragraph{Phrases and Compound Words}
We appended definitions only for single words, which includes both hyphenated and concatenated compound words in German, but did not consider phrases whose translations may not be directly deducible from the constituent words. Similarly, we did not consider separating compound words into the constituent words and recursively searching for definitions if the compound words are not present in the dictionary themselves since subword segmentation often handles these.

\section{Conclusion}\label{conclusion}
In this paper, we proposed using bilingual dictionaries and attention masking to improve translation performance for rare words, a problem that encoder-decoder models continue to struggle with in MT. Our method was to append definitions to source sentences for low-frequency words and use attention masking to associate rare words with their definitions. We found that our method improved MT performance by up to 1.0 BLEU and 1.6 MacroF1. In the future, we are interested in incorporating other external knowledge sources, such as monolingual dictionaries and knowledge graphs, to reduce translation ambiguity and further improve the translation of rare words.

\section{Limitations}\label{limitations}
The following are two limiting factors of our masking approach to including bilingual dictionaries in machine translation: (1) the quality and coverage of the lemmatizer and/or dictionary is a bottleneck to further improvement and (2) appending definitions increases sentence length and therefore runtime.

\section{Acknowledgements}
This material is based upon work supported by the National Science Foundation under Grant No. IIS-2137396.

\begin{small}
\bibliographystyle{apalike-url}
\bibliography{amta2024}
\end{small}

\end{document}